\documentclass[letterpaper]{article} 
\usepackage[preprint]{aaai2027}  
\usepackage[hyphens]{url}  
\usepackage{graphicx} 
\usepackage{natbib}  
\usepackage{caption} 
\usepackage{booktabs}
\usepackage{array}
\usepackage{tabularx}
\usepackage{threeparttable}
\usepackage{multirow}
\usepackage{amsmath,amssymb}
\usepackage{microtype}

\newcolumntype{Y}{>{\raggedright\arraybackslash}X}
\newcolumntype{L}[1]{>{\raggedright\arraybackslash}p{#1}}
\newcolumntype{C}[1]{>{\centering\arraybackslash}p{#1}}

\title{Can Large Language Models Anticipate Behavioral Responses to Social Policies? \\ A Case of Pension Enrollment Prediction among China's Flexible Workers}

\author{
    Yumiao Li\textsuperscript{\rm 1},
    Peixin Liu\textsuperscript{\rm 2},
    Donglin Di\textsuperscript{\rm 3},
    Chen Li\textsuperscript{\rm 1},
    Runhuan Feng\textsuperscript{\rm 1}\corresponding
}
\affiliations{
    \textsuperscript{\rm 1}School of Economics and Management, Tsinghua University, Beijing, China\\
    \textsuperscript{\rm 2}Actuarial Science and Risk Management, University of Illinois Urbana-Champaign, Urbana, IL, USA\\
    \textsuperscript{\rm 3}School of Computing, Harbin Institute of Technology, Harbin, China\\
    fengrh@sem.tsinghua.edu.cn
}

\begin{document}

\maketitle

\begin{abstract}

Assessing the impacts of social policy changes is a widely acknowledged challenge for policymakers. Econometric methods can be unreliable when extrapolating to hypothetical scenarios, while field pilot programs are highly costly. In this paper, we propose using large language models (LLMs) as policy-assessment tools adapted from general-purpose models. We present FlexPension-LLM, the first domain-specialized large language model for a hierarchical pension-enrollment prediction task among flexible workers in China, and introduce DKI-RDistill, which injects policy-grounded cues into the prompt, including Probit-derived marginal effects and hukou-province pension rules. The method then uses LoRA/SFT to distill rationale-augmented supervision into an open-weight MoE student, with teacher errors corrected by regenerating those cases under ground-truth labels. On a CHFS 2019 blind split, FlexPension-LLM achieves 0.9316 Composite F1, surpassing its Claude Sonnet 4.5 teacher and 15 of 17 baselines, and is statistically indistinguishable from Claude Opus 4.6. Across four external surveys, it averages 0.7549 Composite F1 and shows the narrowest performance range among the strongest systems. Component analysis shows that gains come mainly from policy-grounded cue injection and error-filtered supervision, while rationales provide decision traces that can be checked against policy rules.
\end{abstract}

\begin{links}
    \link{Code and reproduction materials}{https://github.com/liym22/FlexPension-LLM}
\end{links}

\section{Introduction}


Understanding the likely impacts of a social policy before it is enacted is essential for sound governance. 
Econometric methods are useful for identifying associations, but they can be unreliable when extrapolating to hypothetical policy scenarios. Large regional pilots are often costly to implement. This motivates models that can simulate individual-level policy responses.

\begin{figure}[t]
\centering
\includegraphics[width=0.95 \linewidth]{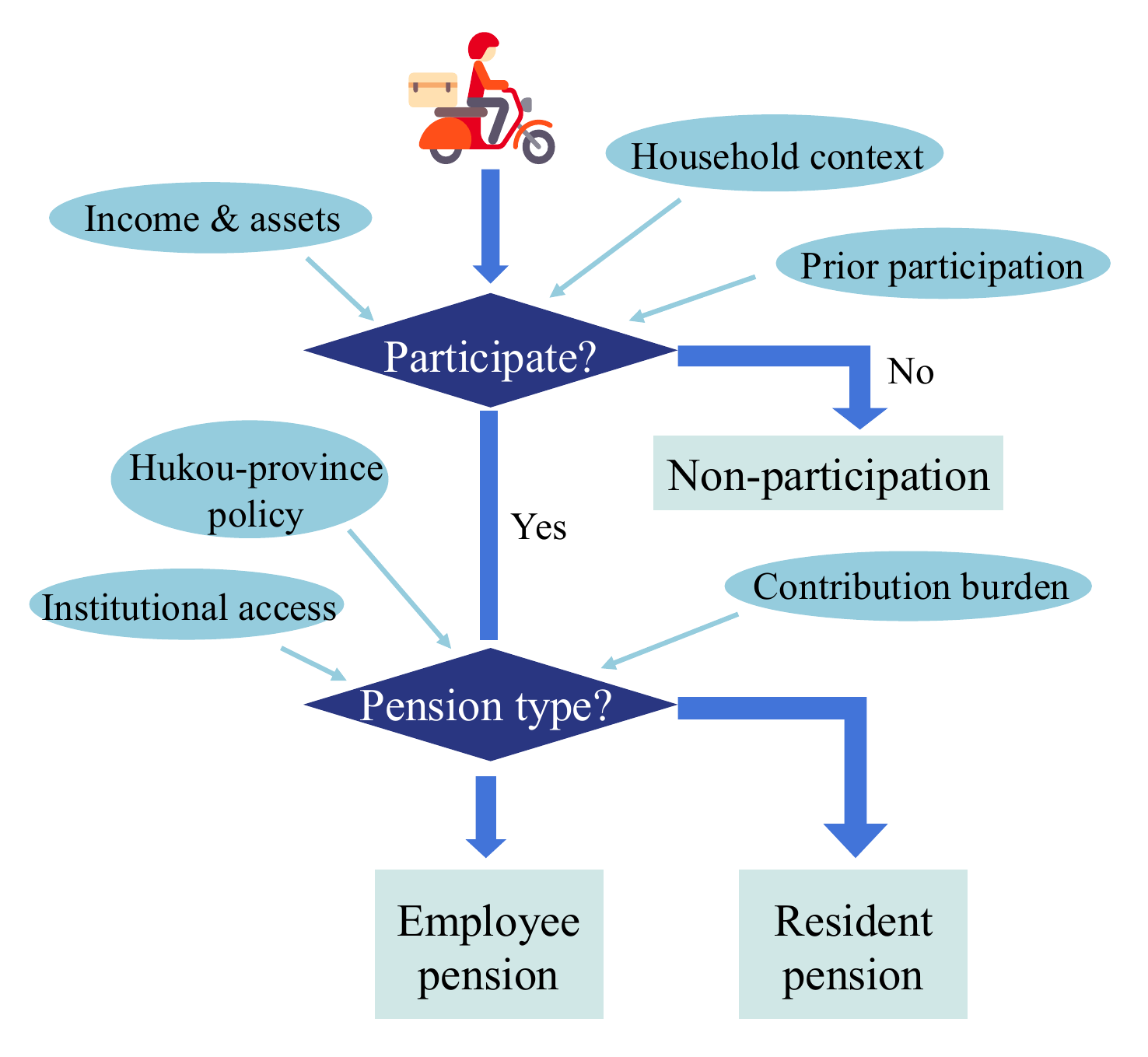}
\caption{Hierarchical pension decision making. Flexible workers first decide whether to participate and then choose between resident- and employee-pension channels. The decision-making process involves multiple factors such as financial status, household context, enrollment history, hukou-province policy, institutional access, and contribution burden.}
\label{fig:problem-overview}
\end{figure}

We instantiate this policy-response problem in flexible workers' pension enrollment in China, where voluntary participation creates a hierarchical decision-making problem summarized in Figure~\ref{fig:problem-overview}. This decision-making process is governed by complex rules and is highly sensitive to marginal policy changes, which is the kind of setting where classical models struggle to capture institutional constraints and behavioral responses. The enrollment setting is large-scale, policy-relevant, and behaviorally open. Recent reports estimate that China's flexible-employment workforce has surpassed 300 million, making it a vital segment of the labor market and an explicit focus of employment policy \cite{bluecollar2026projection,statecouncil2020flexemployment}. 
Prior research documents persistent coverage gaps, heterogeneous pension choices, and welfare losses from enrollment shortfalls and portability frictions \cite{qian2021extension,ma2024precarious,liu2022flexible}.

China's national pension system gives flexible workers two main options. The employment-based Urban Employee Basic Pension is mandatory for formal employees and offers higher benefits, but flexible workers who choose this channel must shoulder the full contribution themselves. The Urban-Rural Resident Pension is more affordable and covers people outside formal employment, but it provides lower benefits. Because participation is voluntary rather than automatically tied to formal employment, the task is not a simple binary classification: a worker first decides whether to participate at all and then, conditional on participation, chooses between resident- and employee-pension channels \cite{yang2022inclusive}.

\begin{figure}[t]
\centering
\includegraphics[width=\linewidth]{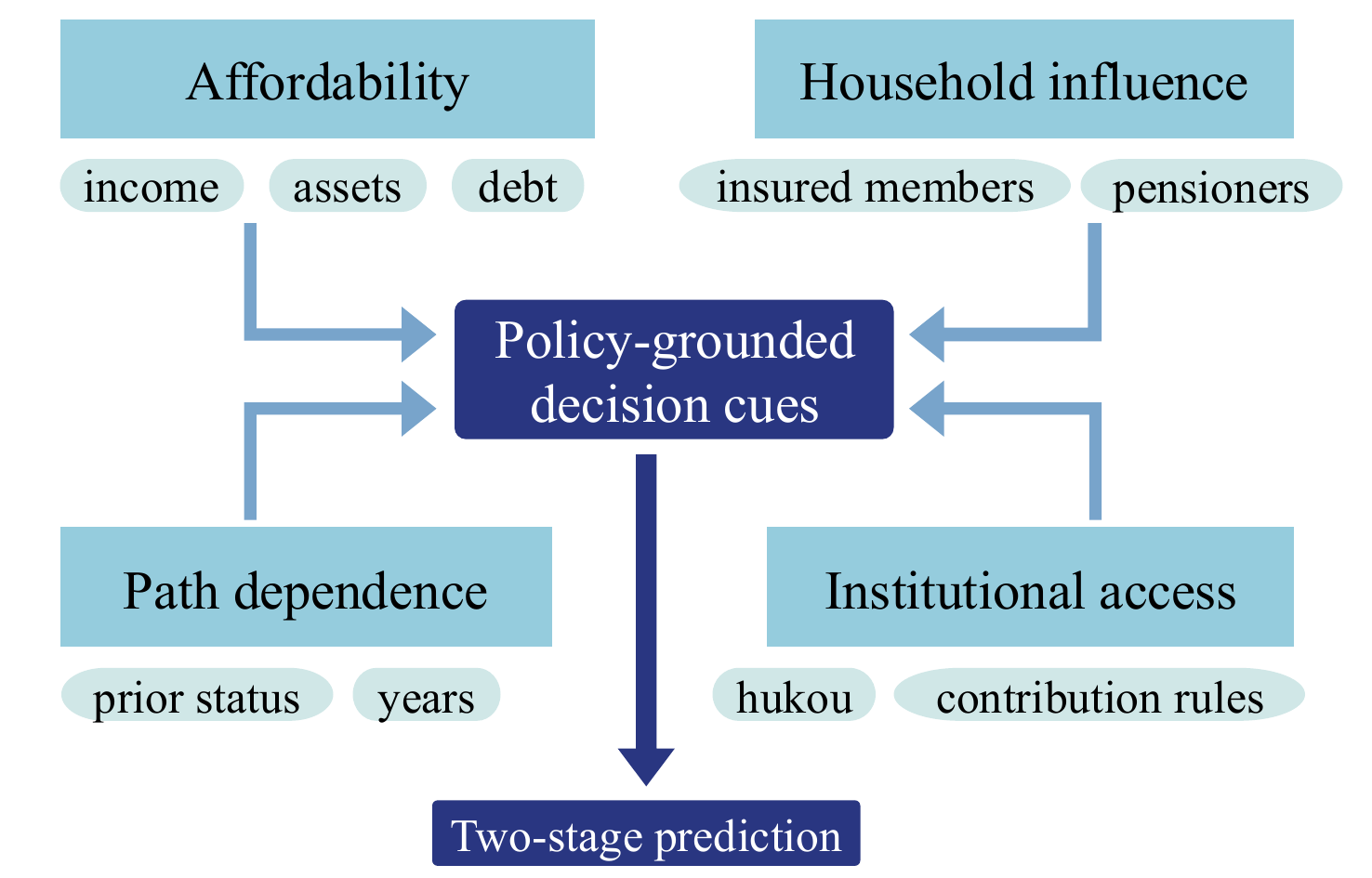}
\caption{Decision signals behind pension-enrollment prediction. Prior social-insurance studies identify economic, household, historical, and institutional signals; we organize these signals into policy-grounded decision cues that support two-stage pension-enrollment prediction.}
\label{fig:decision-signals}
\end{figure}

Predicting this voluntary hierarchical decision requires combining economic, household, historical, and institutional signals. These include personal income, household assets, prior participation, hukou-province policy rules, family pension status, access constraints, and contribution burden. Figure~\ref{fig:decision-signals} organizes these signals into four policy-grounded mechanisms: affordability, household influence, path dependence, and institutional access.


These factors are common targets of econometric analysis \cite{wooldridge2010econometric}, and we use them as empirical priors. However, fitted equations mainly capture population-level associations and are less suited to tracing how income volatility, household support, hukou-linked access, and affordability thresholds combine in a specific case. Standard machine-learning predictors can improve accuracy, but they offer limited explanations tied to policy levers. LLMs offer a different possibility: mapping structured profiles into explicit rationales over behavioral and institutional cues. Yet off-the-shelf LLMs often apply rules superficially, exhibit a rational-actor bias, and show weak consistency in specialized social-science tasks \cite{ziems2024can,manning2024automated}. These limitations motivate a policy-grounded, domain-specialized LLM for pension-enrollment prediction.

We present FlexPension-LLM, the first domain-specialized language model for this hierarchical pension-enrollment prediction task among flexible workers in China. The model takes a survey-grounded worker profile, household context, historical participation information, hukou-province policy parameters, and Probit-derived empirical indicators as input. It outputs both the participation action and the insurance channel, together with a structured rationale. Its intended uses are aggregate prediction, policy simulation, and evaluation of how institutional settings may affect participation decisions.


FlexPension-LLM is adapted through DKI-RDistill. Domain Knowledge Injection converts empirical priors and policy parameters into model-readable cues, including resident- and employee-pension burden ratios and household pension dependency. A strong teacher model then produces structured rationales under this DKI prompt. To improve supervision quality, teacher-correct traces are retained, while teacher-error cases are selectively regenerated under the ground-truth label before LoRA/SFT adaptation.


Using CHFS 2019 as the main dataset, we evaluate FlexPension-LLM on an independent blind test set and four external household-survey datasets. On the blind benchmark, it reaches 0.9316 Composite F1, ranks third among 18 evaluated models, and significantly outperforms 15 of 17 baselines, including Qwen-ZS and the Claude Sonnet 4.5 teacher. Across the four external datasets, it obtains an average Composite F1 of 0.7549, improves over Qwen-ZS and the teacher, and remains statistically comparable to Claude Opus 4.6 and Gemini 3.1 Pro Preview. In a blind expert evaluation with 27 respondents, FlexPension-LLM receives the largest share of trust preferences, 157 of 324 choices (48.5\%).

Our contributions are threefold. First, we introduce FlexPension-LLM, the first domain-specialized language model for this hierarchical pension-enrollment prediction task among flexible workers in China. Second, we introduce DKI-RDistill, a framework that bridges econometric priors, hukou-province pension rules, and error-filtered rationale supervision to produce structured, verifiable decision traces. Third, we validate the model through blind testing, four external datasets, ablations, robustness checks, boundary cases, and expert evaluation, demonstrating near-frontier predictive performance, strong cross-survey generalization, and value for aggregate policy simulation.

\section{Related Work}

Pension and social-insurance participation has been studied with econometric models and, increasingly, machine-learning predictors \cite{wooldridge2010econometric,qian2024firm,mai2024study}. Evidence from China links coverage and pension choices to income, education, hukou status, age, family structure, employer compliance, enforcement, and employment conditions \cite{guo2016pension,qin2015old,qian2021extension}. For flexible and informal workers, participation also depends on contribution thresholds, administrative frictions, perceived benefits, and institutional access. Employee-pension thresholds shape enrollment and welfare stratification, precarious work changes participation incentives, and employment-based versus residency-based pension channels can create different behavioral margins \cite{chu2023thresholds,ma2024precarious,yang2022inclusive}. Policy design, information, and behavioral frictions further affect participation, as shown by evidence on Chengdu's social-insurance design, migrant-worker information interventions, and unclaimed benefits in China's new social pension program \cite{tian2021institutional,giles2021information,chen2020leaving}. These mechanisms make pension enrollment a structured institutional prediction task with multiple behavioral margins.

LLMs have recently been used as economic agents, social simulators, generative agents, and survey-like respondents \cite{argyle2022out,horton2023homo,park2023generative,gao2024large,manning2024automated}. These studies suggest that LLMs can produce plausible behavior and population-conditioned responses, but they also expose reliability concerns: outputs can be sensitive to prompts and demographic conditioning, and models may fail to apply domain-specific rules consistently \cite{ziems2024can}. Our setting makes this limitation central because pension participation requires institutional rule use, affordability calculations, contribution-history reasoning, and household-context interpretation beyond generic social plausibility.

To address this kind of reliability problem, a related technical literature improves LLM behavior through prompt engineering, explicit knowledge injection, chain-of-thought decomposition, self-consistency, tree-structured reasoning, and knowledge distillation \cite{wei2022chain,wang2023self,yao2024tree,hinton2015distilling}. Structured reasoning and teacher-student distillation can transfer task-specific reasoning into smaller models \cite{ho2023large,hsieh2023distilling}, while STaR-style bootstrapping, rationale distillation, and ground-truth-guided regeneration show how generated rationales can become supervision when aligned with labels or reconstructed for difficult cases \cite{zelikman2022star,wu2024radis,cao2026remedi}. Parameter-efficient adaptation methods such as LoRA make this specialization practical for open-source students \cite{hu2022lora,azimi2024kdlora}. FlexPension-LLM combines these ideas with econometric prior extraction and province-level pension-policy cues for a policy-grounded social-science prediction task.

\section{Task and Data}

Building on the decision flow in Figure~\ref{fig:problem-overview}, we formalize pension-enrollment prediction as a hierarchical institutional-decision task. Each instance contains a structured profile of a flexible worker, the worker's household context, historical pension participation, and hukou-province policy context. The label hierarchy has two layers: participation action and, conditional on participation, insurance channel. This gives three terminal labels: non-participation, resident pension insurance, and employee pension insurance. The hierarchy matters because the participation boundary depends on status-quo dependence, household support, and perceived necessity, while the channel decision additionally depends on contribution intensity, subsidy design, affordability, and hukou-linked institutional access.

The task supports aggregate policy simulation and policy evaluation. Model outputs can support counterfactual analysis of how contribution thresholds, subsidies, or local rules may affect participation patterns across heterogeneous worker groups. They are decision-support evidence for aggregate policy analysis, not individual enrollment advice or administrative recommendations.

Our main dataset is CHFS 2019 from the China Household Finance Survey (CHFS) \cite{chfs2019}, from which we construct 15,672 flexible-worker samples: 8,842 participants (56.4\%) and 6,830 non-participants (43.6\%). Table~\ref{tab:task-data-summary} summarizes the label hierarchy, input modules, policy interface, and intended use. Hukou-province policy parameters are manually collected from official provincial or municipal pension-policy documents and matched to each case by hukou province; the supplementary material provides a field preview. For external evaluation, we sample 500 label-stratified cases from each of CHFS 2017, CHIP 2018, CLDS 2018, and CFPS 2018. These large-scale household surveys preserve mappable pension labels and covariates while varying in organization, year, sampling frame, and questionnaire wording, testing transfer beyond the CHFS 2019 construction setting.

\begin{table}[t]
\centering
\small
\setlength{\tabcolsep}{3pt}
\begin{tabularx}{\linewidth}{@{}L{0.26\linewidth}L{0.34\linewidth}Y@{}}
\toprule
Dimension & Specification & Role in prediction \\
\midrule
Main dataset & CHFS 2019; 15,672 flexible-worker samples & Source for model adaptation, blind testing, and empirical priors \\
External datasets & CHFS 2017, CHIP 2018, CLDS 2018, CFPS 2018; 500 sampled cases each & Used only for cross-survey generalization evaluation \\
Target hierarchy & Participation action followed by insurance-channel decision & Aligns metrics with the nested pension-enrollment task \\
Terminal labels & Non-participation, resident pension insurance, employee pension insurance & Final prediction space for evaluation \\
Individual profile & Demographics, employment, income, hukou, mobility, education & Captures worker-level incentives and constraints \\
Household context & Household income, assets, debt, members, pension recipients & Captures family support, demonstration, and substitution effects \\
Participation history & Prior status, contribution years, interruption status & Represents inertia and continuation incentives \\
Hukou-province policy & Resident tiers, employee contribution floors, subsidies, and basic pension parameters & Grounds affordability in local institutional rules \\
Derived indicators & Policy-linked resident-pension and employee-pension burden ratios; household pension dependency & Converts raw survey-policy fields into Probit-informed decision cues \\
Intended use & Aggregate prediction, policy simulation, policy evaluation & Frames outputs as decision-support evidence, not individual advice \\
\bottomrule
\end{tabularx}
\caption{Task and data summary.}
\label{tab:task-data-summary}
\end{table}

\section{Method}

\subsection{Overview of DKI-RDistill}

\begin{figure*}[t]
    \centering
    \includegraphics[width=\textwidth]{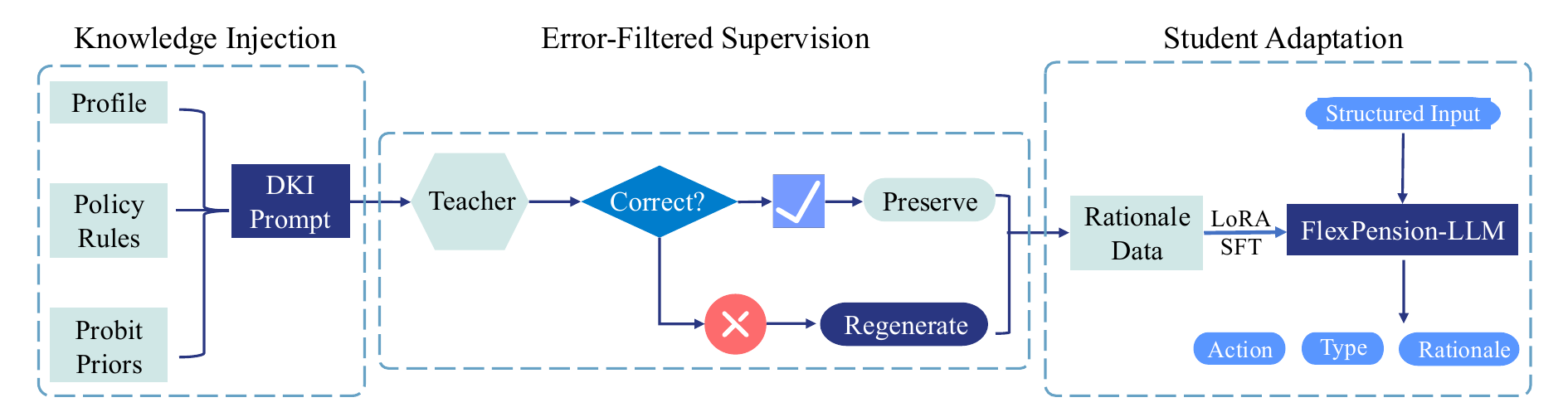}
    \caption{DKI-RDistill workflow. Worker profiles, hukou-province policy rules, and Probit-derived priors are converted into a DKI prompt for teacher generation. Teacher-correct rationales are preserved, teacher-error cases are regenerated under the ground-truth label, and the resulting rationale data are distilled into FlexPension-LLM through LoRA/SFT.}
    \label{fig:framework}
\end{figure*}

Figure~\ref{fig:framework} gives the method workflow. DKI-RDistill first combines worker profiles, hukou-province policy rules, and Probit-derived priors into a policy-grounded DKI prompt. A strong teacher then generates rationale supervision under this prompt: teacher-correct traces are preserved, while teacher-error cases are regenerated under the ground-truth label. Finally, LoRA/SFT adapts the open-weight student into FlexPension-LLM.

The pipeline separates prompt-side cue construction from student adaptation. Econometric analysis and policy rules define which institutional cues should be made explicit; distillation and fine-tuning then convert teacher-guided reasoning traces into stable student predictions.

For example, consider a flexible worker with unstable income, prior resident pension insurance participation, and a hukou province where the minimum resident contribution is affordable but the employee contribution floor is much higher. A flat predictor may overemphasize current income, while the DKI prompt exposes continuation history, household pension context, and policy-linked affordability cues. Under DKI-RDistill, teacher-correct rationales for such cases are preserved, and teacher-error cases are regenerated under the observed label before LoRA/SFT student adaptation.

\subsection{Domain Knowledge Injection}

DKI builds the left side of Figure~\ref{fig:framework}: survey profiles provide case facts, hukou-province policy rules provide local contribution constraints, and Probit estimates identify which mechanisms should be made explicit in the prompt. Model I estimates the participation margin, and Model II estimates employee-pension channel choice conditional on participation. The resulting average marginal effects identify decision-relevant mechanisms: income and assets support participation and employee-channel selection, household enrollment and pension receipt create opposing family effects, contribution years encode path dependence, and hukou/education help explain institutional sorting. The supplementary material maps each mechanism to its empirical basis, injected cue, and prompt role, and reports the full Probit estimates.

DKI turns these mechanisms into three prompt-side indicators: resident-pension burden ratio, employee-pension burden ratio, and household pension dependency. A burden ratio compares the local minimum contribution requirement with the strongest observed capacity signal in the survey profile, so the cue is tied to both household resources and local policy. Let $P_{\min,\text{resident}}$ denote the provincial minimum annual resident contribution, $Y_{\text{ind}}$ individual annual income, $Y_{\text{hh}}$ household annual income, $A_{\text{hh}}$ household assets, $D_{\text{hh}}$ household debt, and $N$ household size. We define
\begin{equation}
R_{\text{resident}} = \frac{P_{\min,\text{resident}}}{\max\left(Y_{\text{ind}}, \frac{Y_{\text{hh}}}{N}, \frac{A_{\text{hh}}-D_{\text{hh}}}{N}\right)}.
\end{equation}
Using the maximum of these capacity signals reduces sensitivity to any single weak or noisy capacity measure.
Analogously, the employee-pension burden ratio is
\begin{equation}
R_{\text{employee}} = \frac{P_{\min,\text{employee}}}{\max\left(Y_{\text{ind}}, \frac{Y_{\text{hh}}}{N}, \frac{A_{\text{hh}}-D_{\text{hh}}}{N}\right)},
\end{equation}
where $P_{\min,\text{employee}}$ is the local minimum annual employee contribution implied by the same policy table. Based on the empirical burden distributions, we use $R_{\text{resident}} < 1\%$ and $R_{\text{employee}} < 25\%$ as practical affordability thresholds; the supplementary material reports the supporting distributions.

Household pension dependency captures whether household resources already rely heavily on pension transfers. We define
\begin{equation}
D_{\text{pension}} = \frac{12 P_{\text{month}}}{Y_{\text{hh}}},
\end{equation}
where $P_{\text{month}}$ is household monthly pension income. Together, these indicators expose affordability, family-pension dependence, and institutional sorting cues before teacher generation begins.

\subsection{Rationale-Augmented Supervision}

The middle branch of Figure~\ref{fig:framework} constructs error-filtered rationale supervision. The DKI prompt is used with Claude Sonnet 4.5 to generate a prediction and rationale for each labeled case. Teacher-correct traces are preserved as usable rationales, and teacher-error cases are regenerated under the ground-truth label so that the student does not imitate an incorrect terminal decision. Let $\mathcal{D} = \{(x_i, y_i)\}_{i=1}^n$ denote the labeled data, $T(\cdot)$ the teacher, and $\hat{y}_i = T(x_i)$ the teacher prediction. We partition the supervision pool into teacher-correct and teacher-error cases:
\begin{equation}
\mathcal{D}_{\text{match}} = \{(x_i, y_i, r_i): \hat{y}_i = y_i\}
\end{equation}
and
\begin{equation}
\mathcal{D}_{\text{regen}} = \{(x_i, y_i, \tilde{r}_i): \hat{y}_i \neq y_i\},
\end{equation}
where $r_i$ is the original teacher rationale and $\tilde{r}_i$ is a regenerated rationale produced under the ground-truth label $y_i$. The final distilled supervision set is
\begin{equation}
\mathcal{D}_{\text{distill}} = \mathcal{D}_{\text{match}} \cup \mathcal{D}_{\text{regen}}.
\end{equation}
In the CHFS 2019 training-and-validation pool, 11,839 samples (88.9\%) are retained as teacher-correct traces and 1,483 samples (11.1\%) are regenerated. This construction keeps usable teacher reasoning and redirects teacher mistakes into label-consistent rationale supervision.

\subsection{Student Adaptation}

The right side of Figure~\ref{fig:framework} internalizes the rationale data into an open-weight student. The student model is Qwen 3.5-35B-A3B, the same unfine-tuned base model evaluated as Qwen-ZS under the DKI prompt. We fine-tune it with LoRA/SFT on the distilled data while keeping the base weights frozen and training only low-rank adapter parameters \cite{hu2022lora}. For input $x_i$, the target sequence $s_i=(r_i,y_i)$ contains both a structured reasoning trace and the final decision label. We minimize the supervised negative log-likelihood:
\begin{equation}
\mathcal{L}(\theta) = -\sum_{(x_i,s_i) \in \mathcal{D}_{\text{distill}}} \sum_{t=1}^{|s_i|} \log p_{\theta}(s_{i,t} \mid s_{i,<t}, x_i).
\end{equation}
Thus the student is trained to output the terminal label and a policy-grounded decision trace anchored to corrected supervision. If the student matches or exceeds the prompted teacher, this indicates the effect of error-filtered supervision and task-specific internalization.

\section{Experimental Setup}

From the 15,672 CHFS 2019 samples, we hold out 2,350 cases as an independent blind test set. The remaining 13,322 cases form the teacher-supervision pool: 11,839 teacher-correct samples are retained directly, while 1,483 teacher-error samples are reconstructed with selective ground-truth-guided rationale regeneration. The resulting distilled data are split into 10,657 training and 2,665 validation samples.

The blind test set is never used in prompt tuning, distilled-supervision construction, fine-tuning, or checkpoint selection. The four external datasets are used only for out-of-domain evaluation and are never incorporated into student fine-tuning. The supplementary material visualizes the full split and supervision flow.

Claude Sonnet 4.5 is selected as the teacher based on a small screening reported in the supplementary material. Qwen 3.5-35B-A3B, an open-weight MoE model, serves as the student base \cite{yang2025qwen3}. Qwen-ZS denotes this unfine-tuned student base evaluated under the DKI prompt, distinct from earlier LoRA batch runs. Correct-only SFT is the label-supervision ablation that excludes regenerated rationale supervision. The final benchmark compares FlexPension-LLM with Qwen-ZS, the Claude Sonnet 4.5 teacher, Correct-only SFT, and recent closed-source baselines including Claude Opus 4.6, Gemini 3.1 Pro Preview, Claude Sonnet 4.6, and DeepSeek V4 Pro.

We report Action F1, Type F1, and Composite F1. Action F1 evaluates whether the individual participates in pension insurance. Type F1 uses the official binary protocol: Urban Employee Pension Insurance is the positive class against other insurance types, and Type F1 is computed only where both the ground truth and prediction indicate participation. Composite F1 is the primary summary metric:
\begin{equation}
\label{eq:composite-f1}
\text{Composite F1} = 0.6 \times \text{Action F1} + 0.4 \times \text{Type F1}.
\end{equation}

For the main tables, we report paired differences with 95\% stratified paired-bootstrap confidence intervals using $B=10{,}000$ resamples and seed 42. The student is fine-tuned with LoRA/SFT and constrained to structured JSON outputs; format-control parameters and the full training configuration are summarized in the supplementary material.

\section{Results and Analysis}

\subsection{Main benchmark}

Table~\ref{tab:main-results} reports the main blind-test benchmark and the external-generalization summary, with deltas computed as FlexPension-LLM minus each baseline. FlexPension-LLM obtains 0.9271 Action F1, 0.9383 Type F1, and 0.9316 Composite F1 on the blind test set, ranking third among 18 evaluated models. Relative to Qwen-ZS, it improves blind Composite F1 by +0.1445 [0.1223, 0.1675]. It also exceeds the Claude Sonnet 4.5 teacher by +0.0107 [0.0005, 0.0212]. Across the full blind benchmark, FlexPension-LLM significantly outperforms 15 of 17 baselines, including Gemini 3.1 Pro Preview, Claude Sonnet 4.6, and DeepSeek V4 Pro; for DeepSeek V4 Pro, the blind delta is +0.0240 [0.0118, 0.0366].

\begin{table*}[t]
\centering
\footnotesize
\setlength{\tabcolsep}{2.5pt}
\begin{threeparttable}
\begin{tabularx}{\textwidth}{@{}L{0.19\textwidth}C{0.085\textwidth}C{0.075\textwidth}C{0.085\textwidth}C{0.075\textwidth}C{0.15\textwidth}C{0.13\textwidth}C{0.13\textwidth}@{}}
\toprule
Model & Blind Action & Blind Type & Blind Comp. & Ext. Avg. & Ext. Range & Blind $\Delta$ & Ext. $\Delta$ \\
\midrule
Qwen-ZS & 0.8378 & 0.7110 & 0.7871 & 0.6378 & 0.5962--0.7264 & +0.1445 & +0.1170 \\
Claude Sonnet 4.5 teacher & 0.9159 & 0.9284 & 0.9209 & 0.6954 & 0.6678--0.7158 & +0.0107 & +0.0595 \\
Correct-only SFT & 0.9258 & \textbf{0.9492} & 0.9352 & 0.7378 & 0.6883--0.7658 & -0.0036 & +0.0171 \\
Claude Opus 4.6 & \textbf{0.9355} & 0.9384 & \textbf{0.9367} & \textbf{0.7724} & 0.7397--0.7953 & -0.0051 & -0.0176 \\
Gemini 3.1 Pro Preview & 0.9081 & 0.9058 & 0.9072 & 0.7631 & 0.7229--0.7951 & +0.0244 & -0.0083 \\
Claude Sonnet 4.6 & 0.9036 & 0.9248 & 0.9121 & 0.7324 & 0.7034--0.7570 & +0.0195 & +0.0225 \\
\textbf{FlexPension-LLM} & 0.9271 & 0.9383 & 0.9316 & 0.7549 & 0.7471--0.7687 & -- & -- \\
\bottomrule
\end{tabularx}
\caption{Main benchmark and external-generalization summary under binary F1 protocols.}
\label{tab:main-results}
\label{tab:external-generalization}
\begin{tablenotes}[flushleft]
\small
\item[] Blind Comp. and Ext. Avg. report Composite F1. Ext. Avg. is the unweighted mean over CHFS 2017, CFPS 2018, CHIP 2018, and CLDS 2018, and Ext. Range gives the min--max across them. Delta columns report FlexPension-LLM minus each baseline; full intervals and the full blind ranking are in the supplementary material.
\end{tablenotes}
\end{threeparttable}
\end{table*}

The blind benchmark places FlexPension-LLM at the near frontier of the evaluated systems. Claude Opus 4.6 has slightly higher Action and Composite F1 point estimates, and Correct-only SFT has the highest Type F1, with both paired intervals crossing zero. This evidence supports a strong near-frontier claim: FlexPension-LLM significantly outperforms most baselines, exceeds its own teacher, and remains statistically tied with the strongest blind-test systems. Correct-only SFT also clarifies the tradeoff: label supervision captures much of the channel boundary, while FlexPension-LLM adds structured rationales exposing affordability, household, history, and hukou cues.

\subsection{Cross-dataset generalization}

Table~\ref{tab:main-results} also summarizes transfer to four household-survey datasets. FlexPension-LLM obtains an external-average Composite F1 of 0.7549, improving over Qwen-ZS by +0.1170 [0.0793, 0.1534] and over the Claude Sonnet 4.5 teacher by +0.0595 [0.0293, 0.0895]. These consistent gains show strong cross-survey transfer beyond the CHFS 2019 construction setting.

Among the strongest external systems, paired differences are small relative to bootstrap uncertainty. Claude Opus 4.6 and Gemini 3.1 Pro Preview have directionally higher external averages, while FlexPension-LLM has higher point estimates than Correct-only SFT and Claude Sonnet 4.6; all four paired intervals cross zero. FlexPension-LLM's range, 0.7471--0.7687, is narrower than Opus 4.6's, supporting a competitive and stable cross-dataset generalization claim.

\subsection{Component analysis}
\label{subsec:components}

\begin{figure}[t]
\centering
\includegraphics[width=\linewidth]{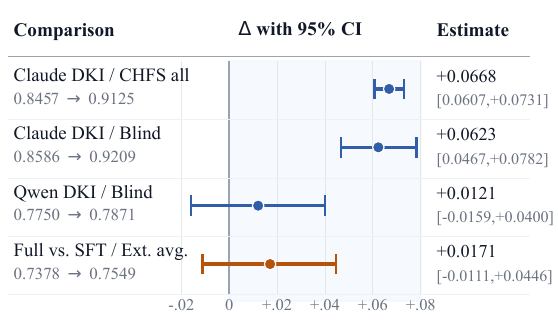}
\caption{Compact component evidence under binary Composite F1. Points show paired deltas, horizontal bars show paired-bootstrap 95\% confidence intervals, and the vertical line marks no gain.}
\label{fig:component-gain}
\end{figure}

Figure~\ref{fig:component-gain} separates the DKI prompt effect from the rationale-supervision effect. The strongest component evidence comes from the teacher side: DKI raises Claude's Composite F1 by +0.0668 [0.0607, 0.0731] on the CHFS 2019 all-sample evaluation and by +0.0623 [0.0467, 0.0782] on the blind subset. Applied directly to Qwen as a prompt-only intervention, DKI gives a small blind-subset gain with a CI crossing zero. This pattern supports the design choice in Figure~\ref{fig:framework}: use DKI to build stronger teacher supervision, then transfer the decision pattern through LoRA/SFT.

Rationale supervision adds a complementary source of value. FlexPension-LLM is higher than Correct-only SFT on the external average (+0.0171), with the paired CI crossing zero, while Correct-only SFT has the higher blind Type F1 in Table~\ref{tab:main-results}. The evidence therefore points to a useful division of labor: label supervision captures much of the channel boundary, and regenerated rationales add inspectable, policy-grounded decision traces over affordability thresholds, household pension dependency, contribution history, and hukou-province access rules.

\subsection{Training efficiency and robustness}

\begin{figure*}[t]
    \centering
    \includegraphics[width=\textwidth]{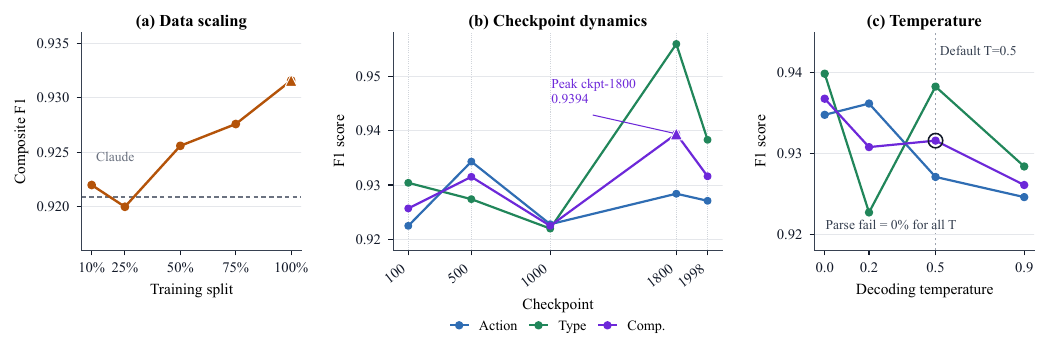}
    \caption{Training efficiency and robustness on the blind test set. Panels show Composite F1 across data scales, checkpoint-level Action/Type/Composite F1, and decoding-temperature stability under constrained JSON output.}
    \label{fig:training-robustness}
\end{figure*}

Figure~\ref{fig:training-robustness} provides auxiliary evidence that the blind-test result is stable across training and decoding choices. Full-data training gives the highest Composite F1 among the tested data scales (0.9316), with a statistically significant but narrow gain over the 25\% setting (+0.0116 [0.0014, 0.0221]). The checkpoint curve enters a high-performance region early and remains stable late in training; checkpoint 1800 has the highest point estimate (0.9394), with its paired comparison against the final checkpoint crossing zero (final minus checkpoint 1800: [-0.0162, 0.0003]). Across decoding temperatures 0.0, 0.2, 0.5, and 0.9, parse success remains 1.0 and all paired comparisons against the default temperature 0.5 cross zero, indicating stable structured-output behavior under the constrained JSON protocol.

\subsection{Error analysis and boundary cases}

\begin{table}[t]
\centering
\footnotesize
\setlength{\tabcolsep}{3pt}
\begin{threeparttable}
\begin{tabularx}{\linewidth}{@{}L{0.30\linewidth}L{0.15\linewidth}L{0.25\linewidth}C{0.18\linewidth}@{}}
\toprule
Comparison & Type & Count change & Rel. change \\
\midrule
Teacher DKI & Total & 468 $\rightarrow$ 246 & $\downarrow\downarrow$ \\
 & Missed & 311 $\rightarrow$ 155 & $\downarrow\downarrow\downarrow$ \\
 & False & 126 $\rightarrow$ 60 & $\downarrow\downarrow\downarrow$ \\
 & Other & 31 $\rightarrow$ 31 & $=$ \\
\addlinespace[1pt]
Student adapt. & Total & 698 $\rightarrow$ 218 & $\downarrow\downarrow\downarrow$ \\
 & Missed & 215 $\rightarrow$ 118 & $\downarrow\downarrow$ \\
 & False & 397 $\rightarrow$ 72 & $\downarrow\downarrow\downarrow$ \\
 & Other & 86 $\rightarrow$ 28 & $\downarrow\downarrow\downarrow$ \\
\addlinespace[1pt]
Teacher$\rightarrow$student & Total & 246 $\rightarrow$ 218 & $\downarrow$ \\
 & Missed & 155 $\rightarrow$ 118 & $\downarrow$ \\
 & False & 60 $\rightarrow$ 72 & $\uparrow$ \\
 & Other & 31 $\rightarrow$ 28 & $\downarrow$ \\
\bottomrule
\end{tabularx}
\caption{Blind-test error transitions.}
\label{tab:error-transitions}
\begin{tablenotes}[flushleft]
\small
\item[] Teacher DKI compares base and DKI prompts; Student adapt. compares Qwen-ZS with FlexPension-LLM; Teacher$\rightarrow$student compares the teacher with FlexPension-LLM. Missed, False, and Other denote missed participation, false participation, and residual/channel mismatches. Arrows mark relative count changes: one for $<25\%$, two for 25--50\%, and three for $>50\%$; upward arrows indicate increases.
\end{tablenotes}
\end{threeparttable}
\end{table}

Table~\ref{tab:error-transitions} shows that teacher-side DKI roughly halves blind-test errors, and student adaptation gives the largest correction relative to Qwen-ZS. FlexPension-LLM sharply reduces the zero-shot student's false-participation and residual/channel errors, while also lowering missed-participation errors relative to the Claude Sonnet 4.5 teacher. The only tradeoff is a modest increase in false-participation errors against the teacher. Overall, the shifts match the DKI-RDistill mechanism: explicit affordability, history, household, and hukou-policy cues reduce false participation and channel confusion, while remaining errors concentrate near the participation boundary.

One boundary case illustrates the benefit of structured cues. Case A has very low annual income, making non-participation plausible from flow-income signals, but it also has substantial household assets, a 0.03\% resident-pension burden ratio, and 14 uninterrupted contribution years. Qwen-ZS predicts non-participation; FlexPension-LLM activates the affordability and contribution-history cues and correctly predicts resident pension insurance participation. Additional boundary cases are reported in the supplementary material.

\subsection{Expert evaluation}

We further ran a blind expert evaluation of rationale quality. 27 respondents with economics, public-administration, insurance, or related backgrounds rated 12 cases, with model identities hidden as Model A, B, and C. As shown in Table~\ref{tab:expert-eval}, FlexPension-LLM received the highest mean soundness and completeness ratings and the largest share of trust preferences, providing expert evidence that DKI-RDistill improves policy-grounded decision rationales.

\begin{table}[t]
\centering
\footnotesize
\setlength{\tabcolsep}{3pt}
\begin{threeparttable}
\begin{tabularx}{\linewidth}{@{}L{0.40\linewidth}ccc@{}}
\toprule
Model & Sound. & Comp. & Trust \\
\midrule
Qwen-ZS & 3.14 & 2.97 & 50 (15.4\%) \\
Claude Sonnet 4.5 teacher & 3.47 & 3.40 & 117 (36.1\%) \\
\textbf{FlexPension-LLM} & \textbf{3.54} & \textbf{3.46} & \textbf{157 (48.5\%)} \\
\bottomrule
\end{tabularx}
\caption{Blind expert evaluation of rationale quality.}
\label{tab:expert-eval}
\begin{tablenotes}[flushleft]
\small
\item[] Sound. and Comp. are mean 1--5 ratings; Trust is the count/share of 324 case-level choices.
\end{tablenotes}
\end{threeparttable}
\end{table}

\section{Conclusion and Limitations}

This paper introduces FlexPension-LLM, the first domain-specialized language model for hierarchical pension-enrollment prediction among flexible workers in China. Its design connects a two-stage policy task, a DKI-RDistill pipeline that injects hukou-province policy rules and Probit-derived priors into rationale supervision, and a broad evaluation across blind testing, external transfer, ablations, robustness checks, boundary cases, and expert ratings. The results show that policy-grounded LLM adaptation can deliver near-frontier predictive performance, stable cross-survey generalization, and expert-preferred decision traces for aggregate policy simulation and evaluation.

The study remains tied to China's pension institutions, household-survey labels, and available policy parameters; transferring it requires rebuilding the policy inputs and validation data. FlexPension-LLM is designed for aggregate policy analysis, not individual enrollment advice, eligibility screening, or automated administrative decisions. Future work should extend the framework to systematic policy counterfactuals and larger expert-in-the-loop policy analysis.

\section{Ethical Statement}

FlexPension-LLM uses structured household-survey records and reports only aggregate, non-identifying results.

\section{Code and Data Availability}

Code and reproduction materials are publicly available at \url{https://github.com/liym22/FlexPension-LLM}. The underlying CHFS, CFPS, CHIP, and CLDS microdata are subject to their respective access conditions and are not redistributed.

\section{Acknowledgments}

The authors gratefully acknowledge the China Household Finance Survey and Research Center at Southwestern University of Finance and Economics for CHFS data, the Institute of Social Science Survey at Peking University for CFPS data, the China Household Income Project (CHIP) team at Beijing Normal University for CHIP data, and the Center for Social Survey at Sun Yat-sen University for CLDS data. The authors are responsible for all analyses, interpretations, and conclusions.

\bibliography{ref/refs}

\clearpage
\setcounter{figure}{0}
\setcounter{table}{0}
\renewcommand{\thefigure}{S\arabic{figure}}
\renewcommand{\thetable}{S\arabic{table}}

\section{Supplementary Material}

This supplementary section reports the data construction flow, supporting tables, prompt excerpts, expert-evaluation details, training diagnostics, and compact examples that complement the main evaluation. It opens with the sample split and evaluation sets so that the subsequent tables can be read against the same data flow.

\subsection{Data Split and Evaluation Sets}

\begin{figure}[h]
\centering
\includegraphics[width=\linewidth]{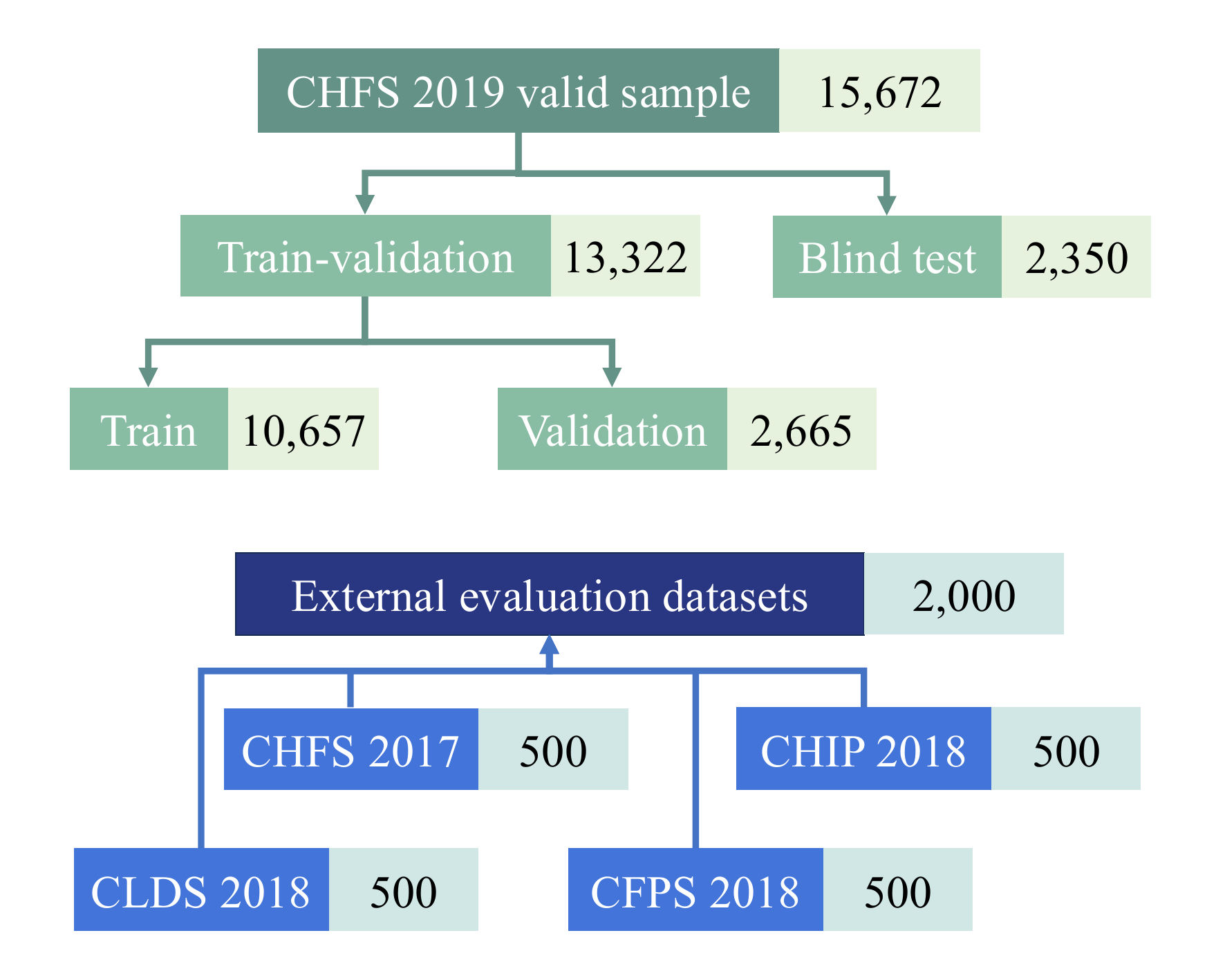}
\caption{Data split and evaluation sets. CHFS 2019 flexible-worker samples are divided into train--validation and isolated blind-test splits. Four additional household-survey datasets, CHFS 2017, CHIP 2018, CLDS 2018, and CFPS 2018, provide 500 sampled cases each for external evaluation.}
\label{fig:data-split}
\end{figure}

Figure~\ref{fig:data-split} summarizes the CHFS split and external evaluation datasets. This diagram is the anchor for the appendix: the screening table explains model-role selection, the DKI prior map explains prompt-side cue construction, and the remaining tables document configuration, prompts, Probit estimates, ablations, and examples.

\subsection{Model Selection and Benchmark Detail}

Table~\ref{tab:model-evidence-appendix} reports the small candidate-model screening used for teacher selection and the full blind-test ranking behind the main benchmark claim.

\begin{table*}[t]
\centering
\footnotesize
\begin{threeparttable}
\setlength{\tabcolsep}{4pt}
\begin{tabularx}{\textwidth}{@{}L{0.32\textwidth}>{\centering\arraybackslash}p{0.16\textwidth}>{\centering\arraybackslash}p{0.16\textwidth}>{\centering\arraybackslash}p{0.11\textwidth}>{\centering\arraybackslash}p{0.17\textwidth}@{}}
\toprule
Model & Screening A-Comp. & Screening V-Comp. & Blind rank & Blind Composite F1 \\
\midrule
Claude Opus 4.6 & / & / & 1 & 0.9367 \\
Correct-only SFT & / & / & 2 & 0.9352 \\
FlexPension-LLM & / & / & 3 & 0.9316 \\
Claude Sonnet 4.5 teacher & 0.8947 & 0.9091 & 4 & 0.9209 \\
Claude Sonnet 4.6 & / & / & 5 & 0.9121 \\
DeepSeek V4 Pro & / & / & 6 & 0.9076 \\
Gemini 3.1 Pro Preview & / & / & 7 & 0.9072 \\
Claude Fable 5 & / & / & 8 & 0.9000 \\
Claude Opus 4.8 & / & / & 9 & 0.8920 \\
Claude Opus 4.7 & / & / & 10 & 0.8823 \\
Qwen3 7 Plus & / & / & 11 & 0.8718 \\
MiniMax M2.5 & / & / & 12 & 0.8663 \\
GPT-5.4 & / & / & 13 & 0.8589 \\
GPT-5.5 & / & / & 14 & 0.8504 \\
GPT-5.4 Mini & / & / & 15 & 0.8463 \\
GPT-5.6 Sol & / & / & 16 & 0.8153 \\
Claude Sonnet 5 & / & / & 17 & 0.8013 \\
Qwen 3.5-35B-A3B / Qwen-ZS & / & / & 18 & 0.7871 \\
\midrule
Gemini 3 Flash & 0.8241 & 0.8866 & / & / \\
MiniMax M2.1 & 0.8132 & 0.8000 & / & / \\
GPT-5 Mini & 0.8053 & 0.8065 & / & / \\
Qwen 3 Max & 0.7496 & 0.7333 & / & / \\
Llama 3.3-70B Instruct & 0.7298 & 0.8571 & / & / \\
DeepSeek V3.2 & 0.6695 & 0.8444 & / & / \\
\bottomrule
\end{tabularx}
\caption{Model-selection and blind-ranking evidence. The table combines the 30-sample candidate screening used for teacher selection with the full blind-test Composite F1 ranking.}
\label{tab:model-evidence-appendix}
\begin{tablenotes}[flushleft]
\small
\item[] A-Comp. averages across runs, and V-Comp. uses vote aggregation. Composite F1 is $0.6 \times$ Action F1 $+ 0.4 \times$ Type F1. `/` indicates that a model was not evaluated in that stage. Blind ranks are based on point estimates; paired-bootstrap uncertainty for key comparisons is reported in the main paper.
\end{tablenotes}
\end{threeparttable}
\end{table*}

\subsection{Method and Implementation Details}

\begin{table*}[t]
\centering
\small
\setlength{\tabcolsep}{4pt}
\begin{threeparttable}
\begin{tabularx}{\textwidth}{@{}L{0.17\textwidth}L{0.25\textwidth}L{0.20\textwidth}Y@{}}
\toprule
Mechanism & Empirical or policy basis & Injected cue & Prompt role \\
\midrule
Historical path dependence & Cumulative contribution years are positively associated with participation and tier persistence. & Prior status, contribution years, interruption status. & Step 1: identify inertia, continuation commitment, and re-evaluation points such as the 15-year threshold. \\
Household demonstration vs substitution & Household enrolled count is positive on the participation margin, while pension recipients/dependency are negative. & Household enrolled members, pension recipients, pension dependency. & Step 2: distinguish family demonstration from pension-substitution incentives. \\
Resident affordability & Resident participants and non-participants have highly overlapping low-burden distributions; medians are 0.09\% and 0.10\%. & $R_{\text{resident}}$ and 1\% affordability threshold. & Step 3: treat resident entry as a low direct-cost barrier unless the burden ratio is unusually high. \\
Employee affordability & Employee participants have a higher burden distribution; median 3.37\%, mean 9.74\%. & $R_{\text{employee}}$ and 25\% affordability threshold. & Step 3: assess whether employee-level contributions are financially feasible under the 25\% threshold. \\
Institutional sorting & Agricultural hukou is negative for employee-pension channel choice, while education is positive. & Hukou, education, mobility, and hukou-province institutional access rules. & Step 4: refine resident-versus-employee channel choice under institutional frictions. \\
Nested decision output & The label has a participation layer and, conditional on participation, an insurance-channel layer. & Structured JSON with reasoning steps and terminal label. & Step 5: aggregate cues into non-participation, resident pension insurance, or employee pension insurance. \\
\bottomrule
\end{tabularx}
\caption{Mapping empirical priors and policy rules into DKI prompt cues.}
\label{tab:dki-prior-map}
\begin{tablenotes}[flushleft]
\small
\item[] Full Probit estimates are reported in Table~\ref{tab:probit-results-appendix}. Burden statistics use 15,634 valid observations (15,672 flexible-worker samples minus 38 with a missing income or asset denominator).
\end{tablenotes}
\end{threeparttable}
\end{table*}

\begin{table}[t]
\centering
\small
\begin{threeparttable}
\begin{tabularx}{\linewidth}{@{}L{0.26\linewidth}L{0.34\linewidth}Y@{}}
\toprule
Dimension & Configuration & Role \\
\midrule
Teacher model & Claude Sonnet 4.5 & Highest Composite F1 in screening \\
Student model & Qwen 3.5-35B-A3B & Open-weight student base used for Qwen-ZS \\
Fine-tuning method & LoRA/SFT & Parameter-efficient low-rank adaptation \\
Hardware & 8$\times$NVIDIA L20 (48 GB) & DeepSpeed ZeRO-3 parallelism \\
Software framework & ms-swift & Native LoRA and DeepSpeed support \\
Training time & $\sim$4.5 hours & Full-data run on the listed hardware \\
Inference temperature & $T = 0.5$ & Default evaluation setting \\
Inference mode & \texttt{enable\_thinking false} & Direct structured JSON output \\
Format-control target & Non-thinking JSON training/inference & Align Qwen training targets with direct structured output \\
Primary metric & Composite F1 & $0.6 \times$ Action F1 $+ 0.4 \times$ Type F1 \\
\bottomrule
\end{tabularx}
\caption{Implementation and format-control summary.}
\label{tab:config-summary-appendix}
\begin{tablenotes}[flushleft]
\small
\item[] Format-control parameters are \texttt{add\_\allowbreak non\_\allowbreak thinking\_\allowbreak prefix} and \texttt{loss\_\allowbreak scale ignore\_\allowbreak empty\_\allowbreak think}; inference uses \texttt{enable\_\allowbreak thinking false}.
\end{tablenotes}
\end{threeparttable}
\end{table}

\begin{table}[t]
\centering
\small
\begin{threeparttable}
\begin{tabularx}{\linewidth}{@{}L{0.34\linewidth}Y@{}}
\toprule
Configuration block & Final value \\
\midrule
LoRA & rank = 32; alpha = 64; dropout = 0.05; target modules = all-linear \\
Optimization & epochs = 3; learning rate = 1e-4; scheduler = cosine; warmup steps = 0; weight decay = 0.1 \\
Batching and precision & global batch size = 16 (1 per device $\times$ 2 grad accumulation $\times$ 8 GPUs); bf16; gradient checkpointing enabled \\
Sequence and distribution & max length = 4096; DeepSpeed ZeRO-3 \\
Randomness and decoding & seed = 42; inference temperature = 0.5 \\
\bottomrule
\end{tabularx}
\caption{Final training configuration used for the full-data LoRA/SFT run.}
\label{tab:training-config-appendix}
\begin{tablenotes}[flushleft]
\small
\item[] These settings are the final values used in the released evaluation pipeline and robustness summaries.
\end{tablenotes}
\end{threeparttable}
\end{table}

\noindent\textbf{Prompt templates.}
Table~\ref{tab:prompt-templates-appendix} compares the Baseline Prompt with the DKI Prompt used for teacher-side inference. The comparison highlights how DKI adds explicit household, affordability, and policy-rule cues while keeping the same structured JSON target.

\begin{table*}[t]
\centering
\footnotesize
\setlength{\tabcolsep}{3pt}
\begin{tabularx}{\textwidth}{@{}L{0.16\textwidth}YY@{}}
\toprule
Component & Baseline Prompt & DKI Prompt \\
\midrule
Role and task & Behavioral-economics and social-security role; infer pension participation and, conditional on participation, insurance channel. & Same task and role, anchored to hukou-province policy rules and Probit-derived decision cues. \\
Input fields & Demographics, hukou and mobility, employment and income, assets and debt, household composition, historical participation, risk preference, and region context. & Baseline fields plus precomputed resident-pension burden ratio, employee-pension burden ratio, household pension dependency, and province-level policy parameters. \\
Reasoning structure & Four steps: historical inertia, liquidity pressure, behavioral/contextual adjustment, and final decision generation. & Five steps: participation-state dependence, household influence, explicit affordability rules, behavioral/contextual adjustment, and final decision generation. \\
Affordability handling & Uses raw financial and household information without explicit threshold cues. & Applies resident-pension burden ratio $<1\%$ and employee-pension burden ratio $<25\%$ as prompt-side affordability cues. \\
Output format & Structured JSON with \texttt{decision\_\allowbreak process.step1-step4} and an \texttt{insurance\_\allowbreak decision} object. & Structured JSON with \texttt{decision\_\allowbreak process.step1-step5} and the same terminal decision object. \\
\bottomrule
\end{tabularx}
\caption{Compact comparison of Baseline and DKI Prompt templates.}
\label{tab:prompt-templates-appendix}
\end{table*}

\medskip
\noindent\textbf{Policy-context field preview.}
To ensure that the prompt-side policy scenario matches province-level institutional details, we construct a policy-parameter table and inject the hukou-province record into each case. Table~\ref{tab:policy-fields-appendix} gives a compact field preview using Beijing as an example.

\begin{table}[t]
\centering
\small
\setlength{\tabcolsep}{4pt}
\begin{threeparttable}
\begin{tabularx}{\linewidth}{@{}L{0.42\linewidth}Y@{}}
\toprule
Field & Example value \\
\midrule
Province & Beijing \\
Average social wage (annual) & 131{,}700 \\
Contribution index range & 0.4 to 1.0 \\
Contribution base range (annual) & 52{,}680 to 131{,}700 \\
Employee contribution rate & 0.2 \\
Employee contribution range (annual) & 10{,}536 to 26{,}340 \\
Resident tier rule & Fixed tiers: 40\%, 60\%, 100\% \\
Resident subsidy detail & [1000,2000): 60; [2000,4000): 90; [4000,6000): 120; [6000,9000]: 150 \\
Basic pension (annual) & 8{,}460 \\
Minimum annual contribution & 940 \\
\bottomrule
\end{tabularx}
\caption{Policy-parameter field preview (Beijing example).}
\label{tab:policy-fields-appendix}
\begin{tablenotes}[flushleft]
\small
\item[] Monetary values are annual RMB amounts unless otherwise indicated; subsidy brackets refer to resident-pension contribution tiers.
\end{tablenotes}
\end{threeparttable}
\end{table}

\begin{table}[t]
\centering
\small
\setlength{\tabcolsep}{3pt}
\begin{threeparttable}
\begin{tabularx}{\linewidth}{@{}L{0.31\linewidth}cY@{}}
\toprule
Variable & Mean / Share & Interpretation \\
\midrule
Pension participation & 56.4\% & participation margin is non-trivial \\
Employee pension & 12.6\% & higher-tier channel is selective \\
Age & 40.03 & mid-career flexible-worker sample \\
Female & 36.3\% & gender composition of the sample \\
Agricultural hukou & 70.0\% & institutional access remains hukou-linked \\
Migrant status & 9.6\% & cross-province mobility is present but limited \\
Individual annual income & 40,602 & direct affordability anchor \\
Household assets & 947,157 & household buffer against contribution burden \\
Historical contribution years & 4.34 & path dependence from prior participation \\
Household insured members & 2.08 & family demonstration effect channel \\
Resident burden ratio & 0.18\% & resident entry is usually affordable \\
Employee burden ratio & 15.51\% & employee entry is more demanding \\
\bottomrule
\end{tabularx}
\caption{Task-relevant descriptive statistics for CHFS 2019.}
\label{tab:descriptive-stats-appendix}
\begin{tablenotes}[flushleft]
\small
\item[] Values are means or shares over the 15,672 CHFS 2019 flexible-worker samples unless otherwise noted. Monetary variables are reported in RMB; burden-ratio means use the 15,634 observations with a valid income or asset denominator.
\end{tablenotes}
\end{threeparttable}
\end{table}

\medskip
\noindent\textbf{Additional supporting artifacts.}
Table~\ref{tab:descriptive-stats-appendix} reports compact descriptive statistics for CHFS 2019, Table~\ref{tab:probit-results-appendix} reports the full Probit estimates behind the DKI prior map, Table~\ref{tab:ablation-generalization-appendix} provides the dataset-level external comparison between FlexPension-LLM and the Correct-only SFT ablation, and Table~\ref{tab:boundary-cases-appendix} summarizes the boundary cases discussed in the main paper.

\begin{table}[t]
\centering
\footnotesize
\setlength{\tabcolsep}{3pt}
\begin{threeparttable}
\begin{tabularx}{\linewidth}{@{}Ycc@{}}
\toprule
Variable & Model I & Model II \\
\midrule
Log individual income & $0.0106^{*}$ & $0.0184^{***}$ \\
Log household assets & $0.0023$ & $0.0278^{***}$ \\
Household enrolled count & $0.1844^{***}$ & $-0.0334^{***}$ \\
Household pension recipients & $-0.1024^{***}$ & $0.0278^{**}$ \\
Household pension dependency & $-0.0000^{**}$ & $0.0000$ \\
Age & $0.0046^{***}$ & $-0.0008$ \\
Education level & $-0.0040$ & $0.0354^{***}$ \\
Agricultural hukou & $0.0438^{***}$ & $-0.1167^{***}$ \\
Cumulative contribution years & $0.0163^{***}$ & $0.0048^{***}$ \\
\midrule
Sample size & 15,642 & 8,825 \\
Pseudo $R^2$ & 0.592 & 0.325 \\
AIC & 8,871.3 & 6,017.9 \\
\bottomrule
\end{tabularx}
\caption{Probit regression results (average marginal effects).}
\label{tab:probit-results-appendix}
\begin{tablenotes}[flushleft]
\small
\item[] Note: $^{***}p < 0.001$, $^{**}p < 0.01$, $^{*}p < 0.05$. Model I: participation decision. Model II: tier choice (employee = 1), estimated on participants only. Sample sizes differ from the full sample because observations with missing covariates in the Probit specification are dropped. The household pension-dependency AME is close to zero after rounding because the variable is measured on its original scale.
\end{tablenotes}
\end{threeparttable}
\end{table}

\begin{figure}[t]
    \centering
    \includegraphics[width=\linewidth]{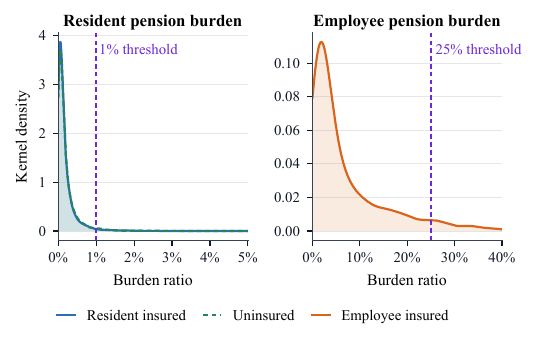}
    \caption{Burden-ratio distributions supporting the affordability thresholds used in DKI. For visual readability, the KDE curves are shown over the main mass of each distribution. Resident-participant and non-participant cases overlap heavily in the low-burden region, whereas the employee burden distribution is shifted to the right.}
    \label{fig:burden-distribution}
\end{figure}

Figure~\ref{fig:burden-distribution} provides the visual distributional check behind the affordability thresholds used in the DKI Prompt.

\begin{table}[t]
\centering
\small
\begin{threeparttable}
\begin{tabular}{@{}lccc@{}}
\toprule
Dataset & FlexPension-LLM & Correct-only & Diff \\
\midrule
CHFS 2017 & 0.7471 & 0.7575 & -0.0104 \\
CFPS 2018 & 0.7522 & 0.7396 & +0.0126 \\
CHIP 2018 & 0.7687 & 0.7658 & +0.0029 \\
CLDS 2018 & 0.7515 & 0.6883 & +0.0632 \\
Average & 0.7549 & 0.7378 & +0.0171 \\
\bottomrule
\end{tabular}
\caption{External Correct-only ablation by dataset under binary Composite F1.}
\label{tab:ablation-generalization-appendix}
\begin{tablenotes}[flushleft]
\small
\item[] Point estimates only; the main paper reports paired-bootstrap confidence intervals for the external-average comparison. Differences on the external average cross zero.
\end{tablenotes}
\end{threeparttable}
\end{table}

\begin{table*}[t]
\centering
\small
\setlength{\tabcolsep}{3pt}
\begin{tabularx}{\textwidth}{@{}L{0.12\textwidth}L{0.24\textwidth}L{0.20\textwidth}L{0.20\textwidth}Y@{}}
\toprule
Case & Key signals & Baseline error & FlexPension-LLM correction & Decision cue \\
\midrule
Case A & 44-year-old female temporary worker; personal annual income 3,000 RMB; household consumption 101,444 RMB; household assets 2.14 million RMB; resident-pension burden ratio 0.03\%; 14 uninterrupted years of resident pension insurance contributions. & Qwen-ZS predicts non-participation by focusing on low income and high consumption. & Predicts resident pension insurance participation, matching the ground truth. & Low burden ratio, asset buffer, and one-year continuation incentive before the 15-year eligibility threshold. \\
Case B & 28-year-old male temporary worker; household income 113,145 RMB; resident-pension burden ratio 0.02\%; two pension recipients; pension dependence 70\%; no prior participation. & Qwen-ZS predicts resident pension insurance participation from apparent affordability. & Predicts non-participation, matching the ground truth. & Pension-dependence signal and non-participation inertia outweigh nominal affordability. \\
\bottomrule
\end{tabularx}
\caption{Boundary cases illustrating corrected participation decisions.}
\label{tab:boundary-cases-appendix}
\end{table*}

\begin{figure*}[t]
    \centering
    \includegraphics[width=0.78\textwidth]{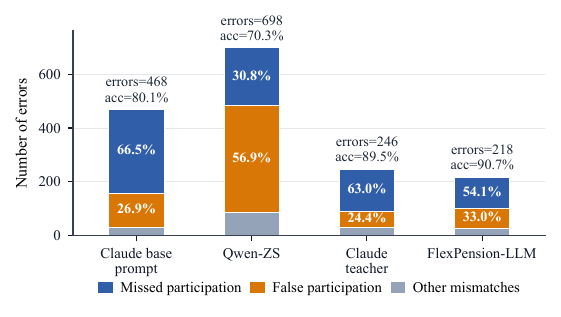}
    \caption{Blind-test error profile for FlexPension-LLM and key baselines. Bars separate missed participation, false participation, and other mismatches; percentages denote each category's share among model errors.}
    \label{fig:appendix-error-profile}
\end{figure*}

Figure~\ref{fig:appendix-error-profile} provides the full error-profile visualization behind the main-paper transition table, including each category's share among model errors.

\subsection{Expert Evaluation Protocol and Results}

We conducted a blind expert evaluation to assess whether the decision rationales produced by FlexPension-LLM are judged as more useful than those from the zero-shot student and the teacher model. 27 respondents completed the study: 15 reported economics backgrounds, five public-administration backgrounds, six insurance or actuarial backgrounds, and one other related background. Their self-reported familiarity with China's pension system was very familiar for nine respondents, familiar for 12, and general for six. The median completion time was 37.7 minutes.

Each respondent evaluated all 12 cases. The cases were sampled from five prediction-outcome strata: three all-correct cases, three cases where only FlexPension-LLM was correct, two cases where only the teacher was correct, two cases where FlexPension-LLM and the teacher were correct but Qwen-ZS was wrong, and two all-wrong cases. Difficult boundary cases were prioritized within the non-unanimous strata. Model identities were hidden as Model A, Model B, and Model C; these correspond to Qwen-ZS, Claude Sonnet 4.5 teacher, and FlexPension-LLM, respectively.

For each model output, respondents rated soundness and completeness on a 1--5 scale. After the true decision was revealed for each case, respondents selected the model explanation they trusted most. Table~\ref{tab:expert-eval-appendix} reports aggregate results only; response metadata such as IP addresses, submission times, and source fields are excluded.

\begin{table*}[t]
\centering
\small
\setlength{\tabcolsep}{4pt}
\begin{threeparttable}
\begin{tabularx}{\textwidth}{@{}L{0.24\textwidth}ccccY@{}}
\toprule
Model & Soundness & Completeness & Mean rating & Trust choices & Respondent-level pattern \\
\midrule
Qwen-ZS & 3.14 (1.04) & 2.97 (1.12) & 3.05 & 50 (15.4\%) & FlexPension-LLM has a higher mean rating for 24 of 27 respondents. \\
Claude Sonnet 4.5 teacher & 3.47 (0.95) & 3.40 (1.01) & 3.43 & 117 (36.1\%) & FlexPension-LLM has a higher mean rating for 15 of 27 respondents and a higher trust share for 17 of 27. \\
\textbf{FlexPension-LLM} & \textbf{3.54 (0.93)} & \textbf{3.46 (0.95)} & \textbf{3.50} & \textbf{157 (48.5\%)} & Highest aggregate soundness, completeness, mean rating, and trust preference. \\
\bottomrule
\end{tabularx}
\caption{Blind expert evaluation of decision-rationale quality.}
\label{tab:expert-eval-appendix}
\begin{tablenotes}[flushleft]
\small
\item[] Soundness and completeness are mean 1--5 Likert ratings with standard deviations in parentheses. Mean rating averages soundness and completeness. Trust choices are counts and shares over 324 case-level selections from 27 respondents and 12 cases. The study used only aggregate summaries in the paper and supplementary material.
\end{tablenotes}
\end{threeparttable}
\end{table*}

\subsection{Training Diagnostics and Supervision Example}

\begin{figure*}[t]
    \centering
    \includegraphics[width=\textwidth]{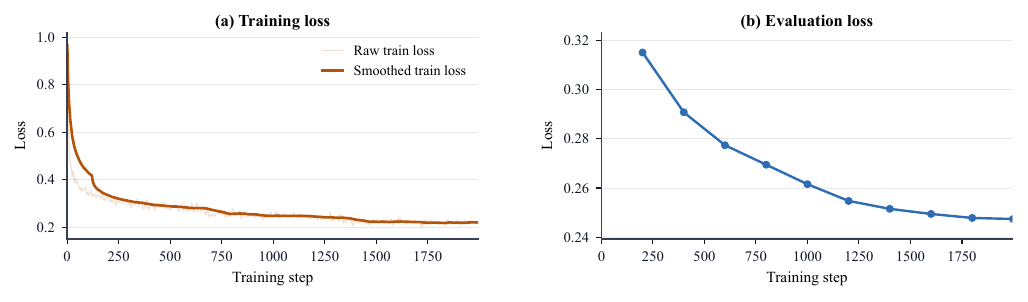}
    \caption{Training and evaluation loss curves from the full-data LoRA run. The training panel shows raw and smoothed training loss, while the evaluation panel reports checkpoint-level validation loss.}
    \label{fig:appendix-loss-curves}
\end{figure*}

Figure~\ref{fig:appendix-loss-curves} adds the optimization view from the full-data LoRA run. The training loss declines rapidly and then stabilizes, while the evaluation loss follows a smoother downward trend with only limited fluctuation. These curves provide an optimization-side complement to the checkpoint-level F1 results.

Table~\ref{tab:jsonl-example} shows a real messages-style JSONL training example in compact form. The user message contains a structured case description with policy context and injected indicators, while the assistant message contains a JSON reasoning-and-decision object used as the fine-tuning target.

\begin{table*}[t]
\centering
\footnotesize
\begin{tabularx}{\textwidth}{@{}L{0.15\textwidth}Y@{}}
\toprule
Field & Example content \\
\midrule
User message (case description) & 28-year-old female; agricultural hukou in Gansu; migrant worker living in Inner Mongolia; annual income 19{,}500 RMB; household income 33{,}135 RMB; household assets 324{,}960 RMB; 8 years of uninterrupted resident pension insurance participation. \\
User message (policy and family cues) & Resident-pension burden ratio: 0.14\%; employee-pension burden ratio: 15.52\%; household pension dependency: N/A; household insured members: 3; household pension recipients: 0. \\
Assistant message (decision trace) & Step 1: path dependence from 8 years of resident participation; Step 2: household demonstration effect; Step 3: resident participation is affordable while employee participation is heavier; Step 4: hukou and mobility favor resident participation; Step 5: maintain resident participation. \\
Assistant message (final decision) & \texttt{action = participate}; \texttt{insurance\_\allowbreak type = resident}; \texttt{annual\_\allowbreak payment = 500}; \texttt{main\_\allowbreak reason = continuation of the 8-year participation path}. \\
\bottomrule
\end{tabularx}
\caption{Compact messages-style JSONL training example.}
\label{tab:jsonl-example}
\end{table*}

The materials above support completeness and reproducibility for the main evaluation while keeping the supplementary file focused on evidence that directly anchors the paper's claims.
\end{document}